\documentclass[11pt]{article}
\usepackage{coling2020}
\usepackage{times}
\usepackage{url}
\usepackage{latexsym}

\usepackage{booktabs}
\usepackage{graphicx}
\usepackage{adjustbox}
\usepackage{comment}
\usepackage{todonotes}
\usepackage{subcaption}
\usepackage{amsmath}
\usepackage{amssymb}
\usepackage[inline]{enumitem}
\usepackage{mathtools}
\usepackage{amsthm}
\DeclareMathOperator*{\argmax}{argmax} 
\newtheorem{theorem}{Theorem}[section]
\newtheorem{corollary}{Corollary}[section]
\theoremstyle{definition}
\newtheorem{definition}{Definition}[section]

\colingfinalcopy 

\title{Mark-Evaluate: Assessing Language Generation using Population Estimation Methods}

\author{Gon\c{c}alo Mordido\\
  Hasso Plattner Institute\\
  Potsdam, Germany\\
  {\tt goncalo.mordido@hpi.de}\\\And
  Christoph Meinel \\
  Hasso Plattner Institute\\
  Potsdam, Germany\\
  {\tt christoph.meinel@hpi.de} \\}

\date{}

\begin{document}
\maketitle

\begin{abstract}
We propose a family of metrics to assess language generation derived from population estimation methods widely used in ecology. More specifically, we use mark-recapture and maximum-likelihood methods that have been applied over the past several decades to estimate the size of closed populations in the wild. We propose three novel metrics: ME$_\text{Petersen}$ and ME$_\text{CAPTURE}$, which retrieve a single-valued assessment, and ME$_\text{Schnabel}$ which returns a double-valued metric to assess the evaluation set in terms of quality and diversity, separately. In synthetic experiments, our family of methods is sensitive to drops in quality and diversity. Moreover, our methods show a higher correlation to human evaluation than existing metrics on several challenging tasks, namely unconditional language generation, machine translation, and text summarization.
\end{abstract}

\section{Introduction}

Population estimation methods have been widely used in ecology to study the development of species over the last several decades~\cite{krebs1989ecological}. Existing population estimation methods focus on open populations, where births, deaths, and migrations are taken into account, or closed population methods, where the population is assumed to remain static over the population estimation study. In this work, we focus on closed population methods and study how their respective population estimates can be used to evaluate an evaluation set of generated samples, given a reference set of real samples. In this work, samples are either contextualized word or sentence embeddings.
We study two mark-recapture methods, namely the Petersen~\cite{petersen} and the Schnabel~\cite{schnabel} estimators, where samples are captured and marked, released, and then recaptured. The number of marked, recaptured, and captured samples can then be used to estimate the population size. We additionally study one maximum-likelihood method, Program CAPTURE
~\cite{capture}, which uses the number of marked or unique samples over multiple captures to estimate the population size.

Accurate evaluation of generated data is essential to correctly measure in what degrees we can improve the overall generation process. Depending on the use case, single-valued metrics may suffice to assess specific conditional language generation tasks, such as machine translation and text summarization, where we are interested in evaluating the similarities of a generated translation or summary to a specific reference translation or summary, respectively. 
On the other hand, on unconditional language generation, for example, it may be useful to have separate measures for the diversity and quality of the generated set, enabling the identification of possible shortcomings of our generation system and try to fix it accordingly. This has been an active area of generative models research, specifically generative adversarial networks~\cite{gans}, where several works have focused on stimulating diversity while maintaining the overall sample quality~\cite{veegan,pacgan,dropoutgan,microbatchgan,sauder2020best}. Hence, depending on the context, a single-valued or double-valued metric may be more desirable.

Mark-Evaluate (ME) is a family of 3 novel language evaluation methods based on the above population estimation methods: ME$_\text{Petersen}$ and ME$_\text{CAPTURE}$ retrieve a single-valued metric to assess an evaluation set, while ME$_\text{Schnabel}$ returns a double-valued metric, separately measuring the quality and diversity of the evaluation set.
Our main contributions can be listed as follows:
\begin{enumerate*}[label=(\roman*)]
    \item Proposal of 3 novel language metrics (Section~\ref{sec:mark_evaluate}) that are sensitive to mode collapse (Section~\ref{sec:mode_collapse}) and quality detriment (Section~\ref{sec:swap}) and show a high correlation to human evaluation on challenging text generation tasks, such as unconditional language generation (Section~\ref{sec:language_generation}), machine translation (Section~\ref{sec:machine_translation}) and text summarization (Section~\ref{sec:text_summarization}).
    \item In-depth study of the language assessment capability of popular existing metrics, \textit{i.e.} FID~\cite{fid}, PRD~\cite{prd} and IMPAR~\cite{impar}, primarily used to evaluate image generation in the past.
    \item Usage of contextual information and different levels of granularity to assess language, by using either contextualized word (Sections~\ref{sec:machine_translation} and ~\ref{sec:text_summarization}) or sentence embeddings (Sections~\ref{sec:synthetic_experiments}, \ref{sec:language_generation}) derived from BERT~\cite{bert}.
    \item Code for the reproducibility of the results will be publicly available.
\end{enumerate*}

\section{Related work}

While acknowledging the importance of traditional evaluation metrics, such as BLEU~\cite{bleu}, ROUGE~\cite{rouge} and METEOR~\cite{meteor}, we will focus on the new trend of unsupervised methods that use embedding representations of pre-trained models to assess a set of evaluation samples. Our family of methods analyzes the data manifold to assess the evaluation set by using \textit{k-nearest neighbors} to determine the capture volume. Several methods have been recently proposed to assess data generation using topological information, however, they were primarily intended to assess image generation~\cite{prd,khrulkov2018geometry,impar,fti}.
In this work, we investigate the performance of such methods in the text domain, analyzing their behavior on synthetic experiments and their correlation with human evaluation.

Precision and recall for distributions (PRD) was proposed by~\newcite{prd} and uses of \textit{k-means}~\cite{kmeans_2} to build histograms of the discrete reference and evaluation distributions over the clusters' centers. The evaluation distribution is then assessed in terms of relative probability densities. Precision is obtained by calculating the probability of an evaluation sample falling within the reference distribution's support. On the other hand, recall is retrieved by calculating the probability of a reference sample falling within the evaluation distribution's support.

\newcite{impar} suggested several improvements to the above method, which we call improved precision and recall (IMPAR). First, instead of \textit{k-means}, they proposed to use \textit{k-nearest neighbors} to approximate the reference and evaluation manifolds by building a hypersphere around each sample to its $k$-th nearest neighbor. Second, they simplify PRD's notions of precision and recall, by calculating the probability of an evaluation sample to fall within at least one reference sample's hypersphere, and vice-versa, respectively. The proposed manifold approximations by the usage of hyperspheres present a simple, yet effective way of representing the reference and evaluation manifold in an explicit, non-parametric way. We build upon this idea and use identical hyperspheres to determine the capture volume used by our different estimators to estimate the population size.

Fréchet Inception Distance or FID~\cite{fid} is a widely used single-valued metric that assesses data similarity by calculating the distance between the reference and evaluation distributions. Even though originally proposed for the image domain, \newcite{semeniuta2018accurate} adapted FID to evaluate text generation by getting vector representations from InferSent~\cite{infersent}, instead of Inception-V3~\cite{inception}. Even though this metric precedes PRD and IMPAR, FID is still commonly used to assess generative models in the image and text domain.

As previously mentioned, we study both the usage of sentence embeddings, derived from SBERT~\cite{sbert}, as well as contextualized word embeddings from BERT~\cite{sbert}, which have been recently shown to improve language assessment, both in a supervised~\cite{bertr,bleurt} and unsupervised manner~\cite{moverscore,bertscore}.
More specifically, BERTScore measures precision and recall from a reference and evaluation text by calculating the required transport of each word of a given text to the most semantically similar word in the other text. On the other hand, MoverScore measures the semantic distance between two texts by calculating the minimum transport required between the reference and evaluation texts. 
These metrics are ideal for conditional text generation, such as machine translation and text summarization, where an evaluation text should match a given reference text.

\section{Mark-Evaluate}
\label{sec:mark_evaluate}

In this work, we consider population estimation methods for closed populations, where the true population size remains constant throughout the estimation study. In our use case, our population consists of two sets, namely a reference set $S_r$ and an evaluation set $S_e$. The true population size ($P$) is then known \textit{a priori} and represents the total number of samples in the two sets: $P=|S_r|+|S_e|$. Given an estimated population size ($\widehat{P}$) from one of the used estimators, we measure the accuracy loss ($A$) as follows:

\begin{equation}
\label{eq:accuracy}
A(P,\widehat{P}) = \text{max}\Big(\Big|\frac{\widehat{P}-P}{P}\Big|,1\Big),
\end{equation}

with a low accuracy loss, \textit{i.e.} $A(P,\widehat{P})\approx 0$, representing a good population estimate, and a high accuracy loss, \textit{i.e.} $A(P,\widehat{P})\approx 1$, otherwise.
Our population estimation methods assume all samples to have an equal chance of capture, which is influenced by our capture volumes: hyperspheres that reaches each reference or evaluation sample's $k$-th nearest reference or evaluation neighbor, respectively.
Hence, if evaluation samples tend not to be inside any reference sample's hypersphere and vice-versa, the population estimate will likely be poor due to the lack of captured samples in the estimation study.

\begin{figure}
\caption{Each sample $s \in S$ and $s' \in S'$ is represented as a blue and red circle, respectively. Marked samples are represented by filled circles. In this illustration, hyperspheres reach to each sample's nearest neighbor of the same set, \textit{i.e.} $K=1$.
For ME$_\text{Petersen}$ and ME$_\text{Schnabel}$, we first capture and mark all samples inside any hypersphere of $s$. Then, for ME$_\text{Petersen}$, we count the number of marked samples or recaptures inside any hypersphere of $s'$. For ME$_\text{Schnabel}$, we perform a similar process iteratively, marking and recapturing samples inside the hypersphere of each $s'$, resulting in all samples being marked in the end.
On the other hand, ME$_\text{CAPTURE}$ captures and marks samples inside each hypersphere of $s$ and $s'$.
}
\centering
\includegraphics[width=0.9\textwidth]{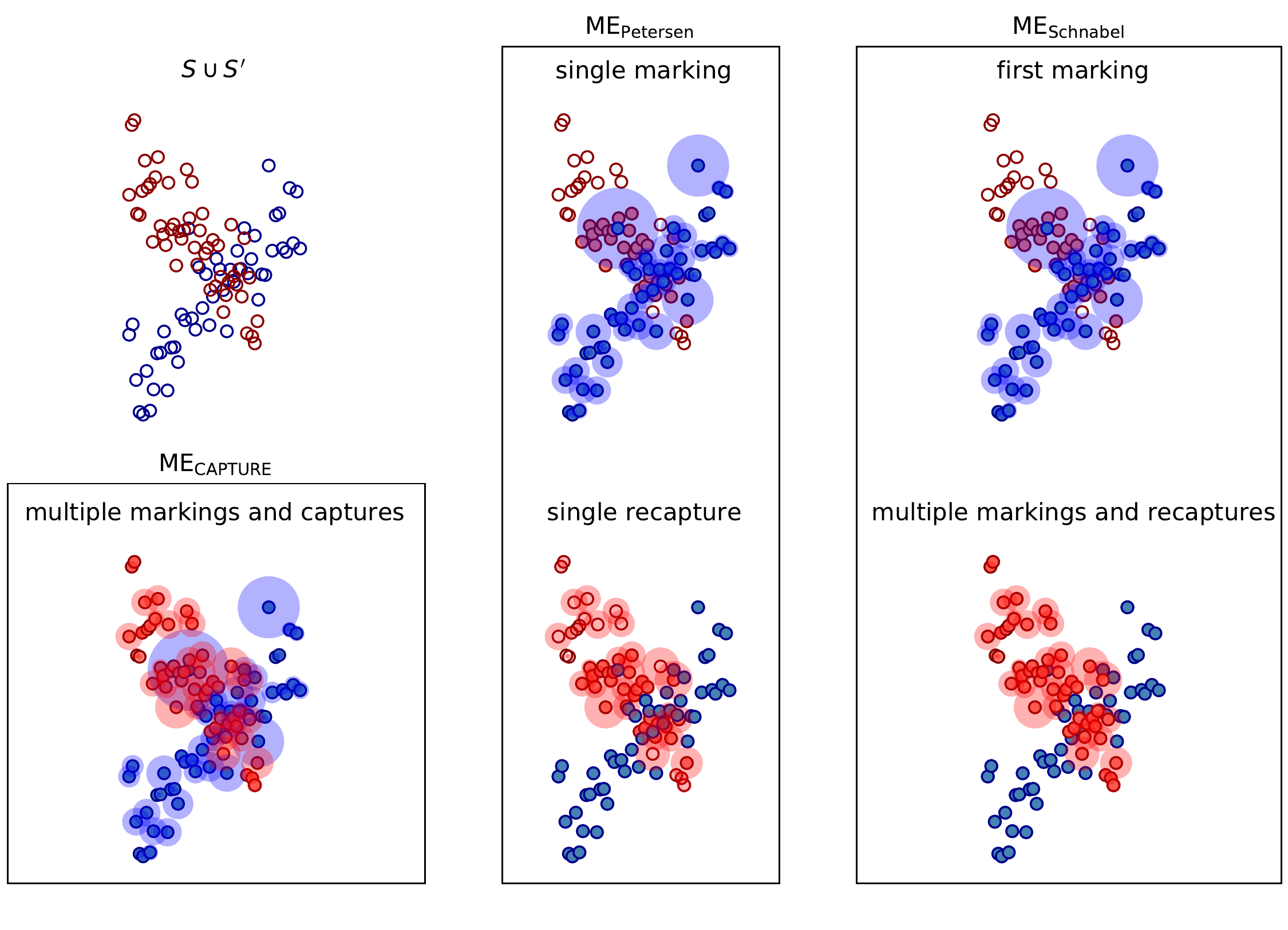}
\label{fig:mark_recapture}
\end{figure}

Our methods can be separated into three categories: single marking and recapture (ME$_\text{Petersen}$), multiple markings and recaptures (ME$_\text{Schnabel}$) and multiple markings and captures (ME$_\text{CAPTURE}$). Let us consider two sets of samples $S$ and $S'$, where each sample is a contextualized word embedding or a sentence embedding derived from BERT, depending on the task. 
Figure~\ref{fig:mark_recapture} illustrates our family of methods. 

Adapting \newcite{impar}'s formulations, we define a binary function $f$ that returns whether a sample $s' \in S'$ lays inside any capture volume or hypersphere of a sample $s \in S$:

\begin{equation}
    f(s', S) = 
    \begin{cases}
        1, \quad \text{if} \quad ||s'-s||_2 \leq ||s - \text{NN}_k(s, S)[-1]||_2 \quad \text{for at least one} \quad s \in S\\
        0, \quad \text{otherwise,}\\
    \end{cases}
\end{equation}

where NN$_k(s, S)$ returns an ordered set containing $s$ and its $k$-nearest neighbors in the set $S$, in ascending order of Euclidean distances to $s$. Hence, NN$_k(s, S)[-1]$ represents the $k$'th nearest neighbor of $s$.
We may refer to individual samples in $S$ and $S'$ as $\{s_1,\ldots,s_{|S|}\}$ and $\{s'_1,\ldots,s'_{|S'|}\}$, respectively.

The Petersen estimator~\cite{petersen}, relies on a single marking step and a single recapture step. It merely assumes that the ratio of marked samples ($M$) in the marking step and the population size ($P$) is equivalent to the ratio of recaptured samples ($R$) and captured samples ($C$) in the recapture step. The population size estimate ($\widehat{P}_\text{Petersen}$) is then calculated as follows:

\begin{equation}
\label{eq:petersen}
    \widehat{P}_\text{Petersen}(S, S') = \dfrac{C(S, S') M(S, S')}{R(S, S')}.
\end{equation}

During the marking step, we mark all samples inside at least one hypersphere of $s$: $M(S, S') = |S| + \sum\limits_{s' \in S'} f(s', S)$. During the recapture step, we do the opposite, marking all samples inside at least one hypersphere of $s'$: $C(S, S') = |S'| + \sum\limits_{s \in S} f(s, S')$. Additionally, in the recapture step, we count the number of captured samples that are already marked from the marking step, \textit{i.e.} the recaptured samples. This corresponds to the number of samples in $S'$ inside at least one hypersphere of $s$ as well as the number of samples in $S$ inside at least one hypersphere of $s'$': $R(S, S') = \sum\limits_{s' \in S'} f(s', S) + \sum\limits_{s \in S} f(s, S')$.

The Petersen estimator was extended by \newcite{schnabel} to incorporate multiple markings and recaptures. The population size estimate ($\widehat{P}_\text{Schnabel}$) is calculated from $T$ consecutive Petersen estimates:

\begin{equation}
\label{eq:schnabel}
    \widehat{P}_\text{Schnabel}(S, S') = \dfrac{C_T(S, S')M_T(S, S')}{R_T(S, S')},
\end{equation}

The set of marked samples at each iteration $t \in \{1,\ldots,T\}$, can be defined recursively as:
\begin{equation}
    M(t, S, S') = 
    \begin{cases}
        S \cup \{s' \in S'| f(s', S) = 1\}, &\text{if }t = 1,\\
        S \cup S', &\text{if }t = T,\\
        M(1,S,S') \cup \Bigg(\bigcup\limits_{i=1}^{t-1}\text{NN}_k(s'_i, S')\Bigg), &\text{otherwise}.
    \end{cases}
\end{equation}

ME$_\text{Schnabel}$'s first marking step is identical to ME$_\text{Petersen}$'s single marking step ($M(1,S,S')=M(S,S')$), with all samples in $S$ as well as samples in $S'$ that are inside at least one hypersphere of $s$ being marked. By the final marking step, all samples will be marked since we iterated through all of them: $M_T(S,S') = |M(T,S,S')|$. For the other iterations, $1<t<T$, samples in $S'$ that are captured, \textit{i.e.} are $k$-nearest neighbors of the $s'$ being iterated, but are not yet marked, are added to the marked set.

After all recapture steps, which excludes the first marking step, the number of captured samples will be the number of samples in $S'$ and their respective $k$'th nearest neighbors as well as samples in $S$ that are inside the hypersphere of each $s'$: $C_T = (K+1)*|S'| + \sum\limits_{s' \in S'} \sum\limits_{s \in S} f(s, \text{NN}_k(s', S'))$.
Since all samples in $S$ have been marked in the first marking step, the number of total recaptures is the number of samples in $S$ inside the hypersphere of each $s'$ as well as the number of $k$-nearest neighbors of the iterated $s'$ that have already been marked: $R_T(S,S') = \sum\limits_{i = 1}^{|S'|} \sum\limits_{j = 1}^{|S|} \Big(f(s_j, \text{NN}_k(s'_i, S')) + |M(i,S,S') \cap \text{NN}_k(s'_i, S')|\Big)$. 

Both ME$_\text{Petersen}$ and ME$_\text{Schnabel}$ are mark-recapture methods since they rely on marking and recapturing information to estimate the population size. We further used a maximum log-likelihood method: the model null from Program CAPTURE~\cite{capture}. By considering the total number of marked samples ($M_T$) and the total number of captures ($C_{total}$) over $T$ iterations, with $T=|S \cup S'|$, we iterate through several provisional population estimates ($P_\text{CAPTURE} \in \mathbb{N}_{\geq M}$) and compute their log-likelihood:

\begin{equation}
\label{eq:capture}
\begin{split}
    L_n(P_\text{CAPTURE};S,S') = \ln\Bigg(\dfrac{P_\text{CAPTURE}!}{(P_\text{CAPTURE}-M_T(S,S'))!}\Bigg) + C_{total}(S,S') \times ln\Big(C_{total}(S,S')\Big)\\ + \Big(TP_\text{CAPTURE}-C_{total}(S,S')\Big) \times \ln\Big(TP_\text{CAPTURE}-C_{total}(S,S')\Big) - (TP_\text{CAPTURE})\ln(TP_\text{CAPTURE}).
\end{split}
\end{equation}

The total number of captures corresponds to the number of samples in $S$ and $S'$ and their respective neighbors, as well as the number of samples in $S$ inside the hypersphere of a given $s'$ and vice-versa:
$C_{total}(S,S') = \sum\limits_{s \in S} \sum\limits_{s' \in S'} ( f(s', \text{NN}_k(s, S)) + |\text{NN}_k(s, S)| ) + \sum\limits_{s' \in S'} \sum\limits_{s \in S} ( f(s, \text{NN}_k(s', S')) + |\text{NN}_k(s', S')|)$. 
The final population estimate ($\widehat{P}_\text{CAPTURE}$) is then the estimate that maximizes Equation~\ref{eq:capture}: 

\begin{equation}
\label{eq:capture_2}
\widehat{P}_\text{CAPTURE}(S,S') = \argmax\limits_{P_\text{CAPTURE}}\widehat{L_n}(P_\text{CAPTURE};S,S').
\end{equation}

Our family of methods uses the accuracy loss of each estimator to compute their scores as follows:

\begin{equation}
\label{eq:mark_evaluate}
    \text{ME}_\text{\{Petersen, Schnabel, CAPTURE\}}(S,S') = 1 - A\Big(P,\widehat{P}_\text{\{Petersen, Schnabel, CAPTURE\}}(S,S')\Big).
\end{equation}

Note that, due to its iterative nature, ME$_\text{Schnable}$ may be used to separately assess the quality and diversity of an evaluation set $S_e$ given a reference set $S_r$. More specifically, quality may be calculated by ME$_\text{Schnable}(S_r, S_e)$, whereas diversity may be measured by ME$_\text{Schnable}(S_e, S_r)$. On the other hand, ME$_\text{Petersen}$ and ME$_\text{CAPTURE}$ are single-valued metrics, since ME$_\text{Petersen}(S_r,S_e)$ = ME$_\text{Petersen}(S_e,S_r)$ and ME$_\text{CAPTURE}(S_r,S_e)$ = ME$_\text{CAPTURE}(S_e,S_r)$. 
We refer to the Appendix for theoretical discussions. 

\begin{figure} 
\caption{Effects of using different capture volumes, \textit{i.e.} changing the number of neighbors ($K$), in the population size estimated of each estimator (left) and our respective method's score (right).}
\centering
\includegraphics[width=1.0\textwidth]{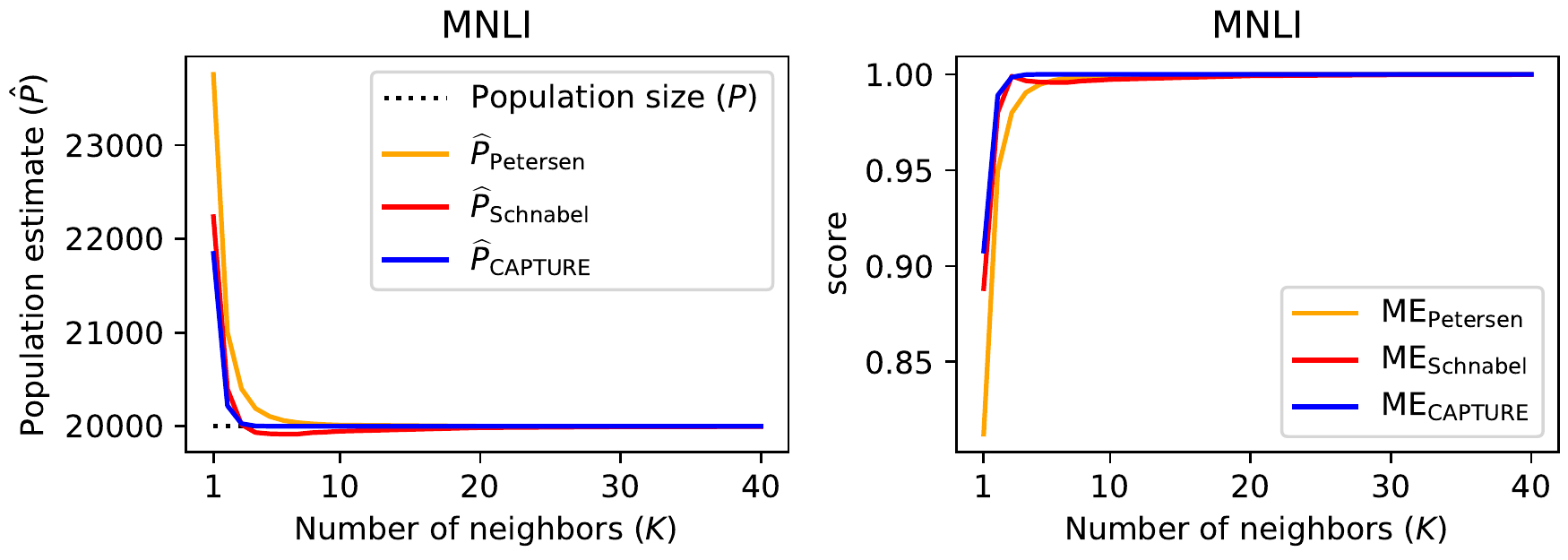}
\label{fig:num_k}
\end{figure}

To study the effects of different capture volumes, determined by different $K$, we used SBERT to get the sentence embeddings of 10k training sentences from MNLI~\cite{williams2017broad} as the reference set, and 10k validation sentences as the evaluation set. Results are shown in Figure~\ref{fig:num_k}, with $K \in \{1,\ldots, 40\}$. We observe that as $K$ increases, the population size estimated by all estimators converges to the true population size. In turn, the scores of our family of methods also converge to their maximum value of 1.

\section{Synthetic experiments}
\label{sec:synthetic_experiments}

\begin{figure}
\caption{Mode collapse experiment where sentences from certain topics are dropped from the evaluation set. Quality (full lines) is expected to remain constant while diversity (dotted lines) is expected to drop as mode collapse aggravates.}
\centering
\includegraphics[width=1.0\textwidth]{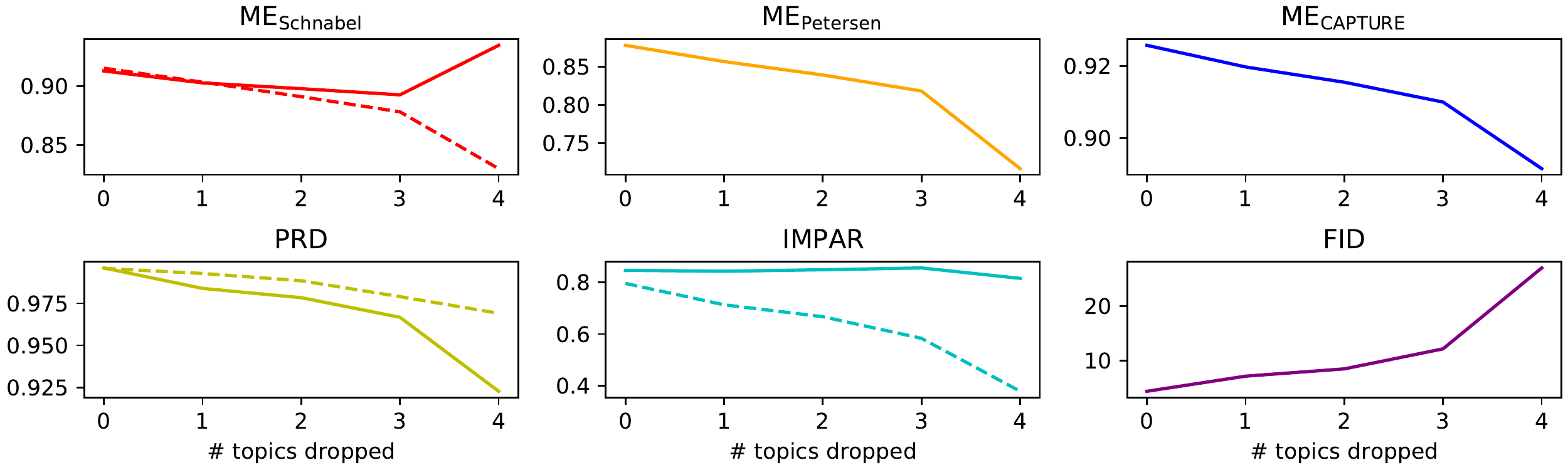}
\label{fig:mode_collapse}
\end{figure}

To simulate drops in quality and diversity, we used the MNLI dataset~\cite{williams2017broad}, which consists of 433k sentence pairs annotated with one out of 5 possible topics. 
Since we were interested in the sentence-pair information for these experiments, we treated each sentence independently. Our reference set consists of sentences from the training set, whereas our evaluation set has sentences from the validation set. We kept the size of the reference and evaluation sets equal throughout our experiments to reduce possible method instabilities regarding sample size. We follow the experiments in \newcite{semeniuta2018accurate} and simulate diversity loss by dropping sentences from certain topics (Section~\ref{sec:mode_collapse}), whereas quality detriment is induced by swapping the words of each sentence (Section~\ref{sec:swap}). For both experiments, we use SBERT embeddings from BERT-base pre-trained on SNLI
~\cite{snli} and MNLI~\cite{williams2017broad} datasets ( \textit{'bert-base-nli-mean-tokens'~\footnote{https://github.com/UKPLab/sentence-transformers}}).

\subsection{Mode collapse}
\label{sec:mode_collapse}

To evaluate mode collapse, we dropped the sentences from specific topics from the evaluation set, containing sentences from all the available topics. Thus, the evaluation set only contains sentences from a subset of topics. What differs at each step is the number of topics included in the evaluation set: for example, dropping one topic means that the evaluation set only contains samples from the rest of the four available topics. The reference set remained unaltered throughout this process. We used 4k reference and 4k evaluation samples throughout this experiment.

We expect quality assessments to remain constant and diversity assessments to drop as fewer topics are represented in the evaluation set. For single-metric methods, we expect a detriment of the overall score throughout the mode dropping process. Results are presented in Figure~\ref{fig:mode_collapse}, where we observe that our family of methods displays the expected behavior. IMPAR and FID also show expected performance (note that higher FID is worse since it represents a distance from the reference and evaluation distributions). On the other hand, PRD's quality assessment or precision drops significantly as mode collapse aggravates, which is not expected since the quality of the evaluation set is not affected in this experiment.
We further observe that all methods show high sensitivity when only one topic is represented in the evaluation set. For example, when 4 topics are dropped, IMPAR's quality assessment shifts by $\approx 0.4$, while ME$_\text{Schnabel}$ and ME$_\text{Schnabel}$ shifts by $\approx 0.5$. Hence, both methods show similar variance, despite the visualization contrast originated from different y-scales.

\subsection{Word swap}
\label{sec:swap}

To evaluate quality detriment, we swapped the words of each sentence in the evaluation set with a certain swap probability. Similarly to the mode collapse experiment, the reference set remains constant throughout this study. We used 10k reference and 10k evaluation samples. 
\newcite{semeniuta2018accurate} showed that sentence embeddings derived from InferSent~\cite{infersent} and Transformers~\cite{vaswani2017attention} models were unable to detect similar quality perturbations. However, we observe that BERT embeddings can capture such quality detriment, observed by the variance of the scores of all the tested methods.

\begin{figure} 
\caption{Swap experiment where words from each sentence in the evaluation set are swapped with a certain swap probability. Quality (full lines) is expected to drop while diversity (dotted lines) is expected to remain constant as the swap probability increases.}
\centering
\includegraphics[width=1.0\textwidth]{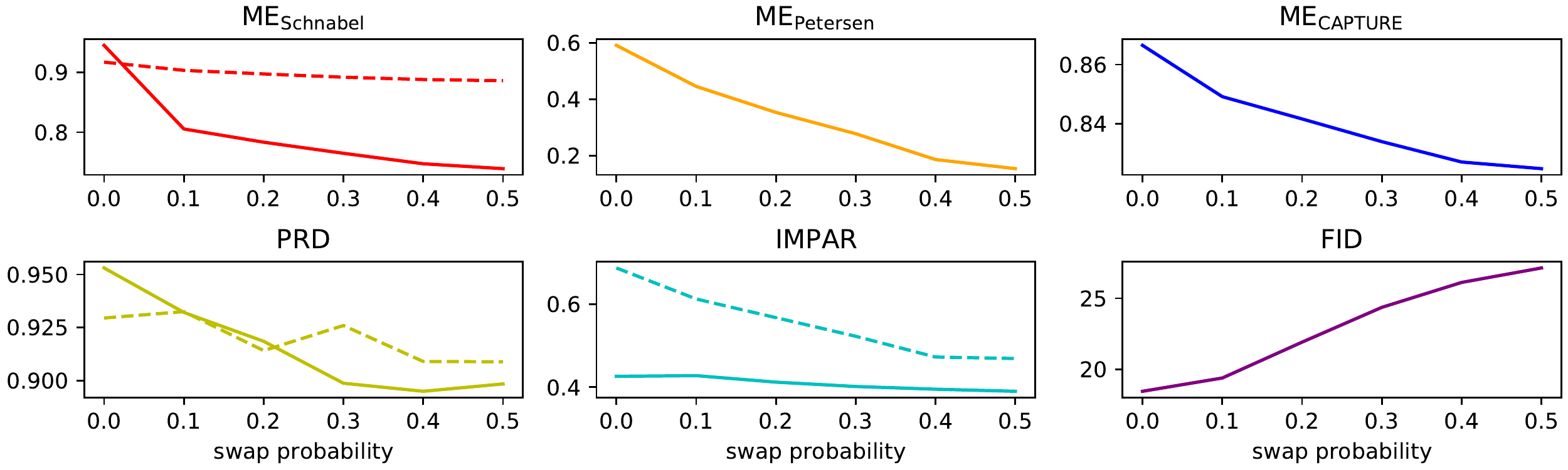}
\label{fig:swap}
\end{figure}

For this experiment, precision is expected to drop, while recall should remain constant. The overall score of single-valued metrics should deteriorate as the swap probability increases. Results are presented in Figure~\ref{fig:swap}. Both the quality and diversity assessments of ME$_\text{Schnabel}$ show the expected behavior, similarly to our single-metrics and FID. On the other hand, the recall or diversity assessment of both PRD and IMPAR drops unexpectedly. Moreover, IMPAR's quality or precision does not drop as significantly at higher swap probabilities, which is not desirable. 

\section{Language generation}
\label{sec:language_generation}

We further assessed the text generated by ten different language generation models presented in \newcite{eval_all}. The models include a traditional language model and several types of autoencoders, namely variational, adversarial, adversarially regularized, and plain autoencoders. We used the human ratings assigned to each model's fluency presented in their work to study the correlation of our family of methods and other tested metrics to human evaluation. 
Reverse and forward cross-entropy, \textit{i.e.} Reverse CE and Forward CE, have been commonly used to assess text generation in the past~\cite{eval_all,semeniuta2018accurate,lm_ppl}. 
The reported Reverse CE and Forward CE results were taken from \newcite{eval_all}, obtained by training a language model on English Gigaword~\cite{napoles2012annotated}. 
We refer to \newcite{eval_all} for additional details.

\begin{table}[b] 
\vskip 0.15in
\begin{center}
\begin{small}
\begin{sc}
\centering
\begin{tabular}{ccccccccc}
\toprule
{\bf Correlations} & Forward CE & Reverse CE & {FID} & {PRD} & {IMPAR} & {\bf ME$_\text{Schnabel}$} & {\bf ME$_\text{Petersen}$} & {\bf ME$_\text{CAPTURE}$} \\
\midrule
\bf Pearson $r$ & 0.606 & 0.440 & 0.902 & 0.830 & 0.745 & \bf 0.917 & 0.872 & \textbf{0.902}\\
\bf Kendall $k$ & 0.556 & 0.333 & 0.867 & 0.822 & 0.778 & \bf \underline{0.911} & \bf \underline{0.911} & \bf 0.867\\
\bf Spearman $p$ & 0.697 & 0.491 & 0.964 & 0.939 & 0.903  & \bf \underline{0.976} & \bf \underline{0.976} & \bf 0.964\\
\bottomrule
\end{tabular}
\end{sc}
\end{small}
\end{center}
\caption{Absolute correlations to human evaluation regarding the fluency of 10 different models with the best $K$. Best scores for each correlation are underlined. Bold values represent the correlations where our methods outperform or match all of the other methods' performance. 
}
\label{tab:human_eval_correlations_best}
\end{table}

We used SBERT embeddings from BERT-large trained on SNLI and MNLI datasets (\textit{'bert-large-nli-mean-tokens'}) since they achieved the best-reported performance in
~\newcite{sbert}. Note that, since human evaluation is only related to each model's fluency, we only report the quality assessment scores for ME$_\text{Schnabel}$, PRD, and IMPAR.
For our family of methods, as well as PRD and IMPAR, we iterate through $K$ values until correlation drops and present the results with the best $K$ of each method. Table~\ref{tab:human_eval_correlations_best} shows the Pearson $p$, Kendall $k$, and Spearman $p$ correlations to human evaluation. 

Overall, our family of methods achieves the highest correlations to human evaluation. Note that despite being outperformed by FID, ME$_\text{Petersen}$ still outperforms PRD and IMPAR across all correlations. Additional results with default $K$ for our family of methods, PRD, and IMPAR, as well as a comparison with InferSent and different SBERT embeddings, are provided in the Appendix.

\section{Contextualized word embeddings}
\label{sec:word_embeddings_mt}

We will now shift our focus to conditional language generation under finer-grained representations, \textit{i.e.} contextualized word embeddings. We used embeddings from BERT-base fine-tuned on MNLI, identically to~\newcite{moverscore}. For a fair comparison, we used the same embedding representations for all the methods in the following experiments. Due to the likely imbalance of reference and evaluation samples, we only report the quality assessment or precision of double-valued metrics. Similarly to Section
~\ref{sec:language_generation}, we report the results with the best $K$. Additional results with default $K$ can be found in the Appendix.

\begin{figure} 
\caption{When using contextualized word embeddings in our family of methods, each word is represented by five samples corresponding to the embeddings of the last five layers of BERT. We used the same procedure as Mark-Evaluate for PRD and IMPAR. On the other hand, MoverScore uses p-means to aggregate the information of the last five embeddings, and BERTScore uses the embeddings of a given layer for each word.}
\centering
\includegraphics[width=1.0\textwidth]{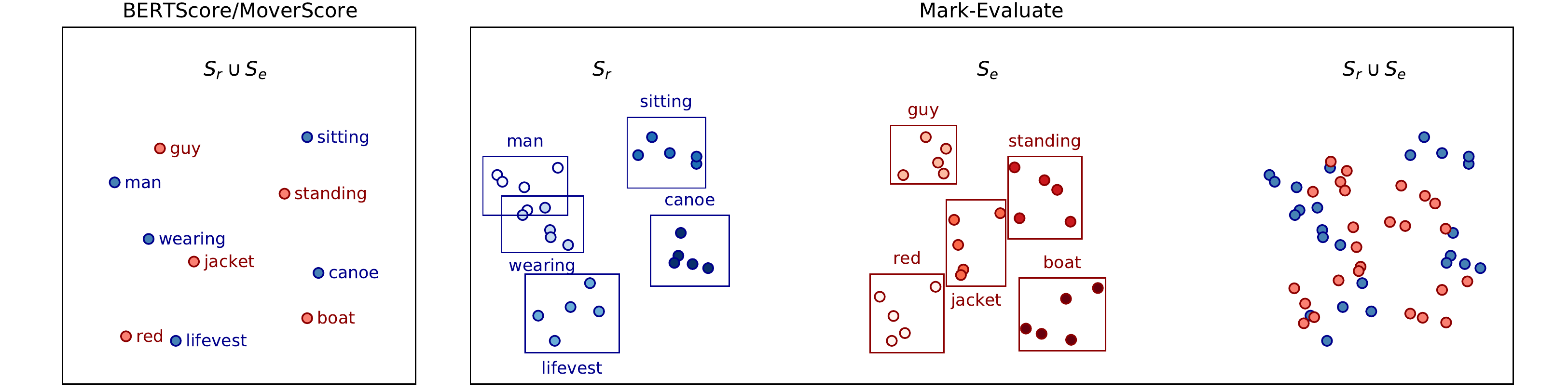}
\label{fig:word_embeddings}
\end{figure}

Using the information of the last layers of BERT has been shown to help in several downstream tasks~\cite{liu_last_layers}. This has also been shown for language assessment, observed by, for example, the fact that the best performing layers of BERTScore are often latter layers~\cite{bertscore}. MoverScore extends this thinking and aggregates the representations of the last five layers of BERT with $p$-means.
For our methods, instead of aggregating or routing this information, we use the vector representation from the last five layers for each specific word. Thus, each word has five representations, defined as five samples, in our scheme. See Figure~\ref{fig:word_embeddings} for an illustration of this process. This also allows us to produce a better population estimate in the end due to the increase of the sample size.

\subsection{Machine translation}
\label{sec:machine_translation}

We start by assessing system-level machine translations from the WMT17 metrics task~\cite{wmt17}. We evaluated the different methods on the five language pairs provided by \newcite{moverscore}'s implementation \footnote{https://github.com/AIPHES/emnlp19-moverscore}. Namely, we assess translations from Czech (cs), German (de), Russian (ru), Turkish (tr), and Chinese (zh) to English (en). Each language pair has around 3k reference with the respective evaluation translations from multiple systems (the number of systems for each language pair varies).

\begin{table*}
  \centering
  \begin{adjustbox}{width=1.0\textwidth}
  \begin{tabular}{ccccccccccccccccc}
    \hline
    \textbf{Translations} & BERTScore & MoverScore & PRD & IMPAR & \textbf{ME$_\text{Schnabel}$}  & \textbf{ME$_\text{Petersen}$} & \textbf{ME$_\text{CAPTURE}$}\\
    \hline
    \bf cs-en ($r$) & 0.966 & 0.983 & \underline{0.992} & 0.987 & 0.989 & 0.988 & 0.987\\ \hline
    \bf de-en ($r$) & 0.859 & 0.920 & 0.769 & 0.934 & \bf 0.944 & \bf \underline{0.953} & \bf \underline{0.953}\\ \hline
    \bf ru-en ($r$) & 0.868 & 0.921 & \underline{0.933} & 0.896 & 0.902 & 0.908 & 0.908\\ \hline
    \bf tr-en ($r$) & 0.938 & 0.931 & 0.935 & 0.959 & \bf \underline{0.970} & \bf 0.960 & \bf 0.959\\ \hline
    \bf zh-en ($r$) & 0.894 & 0.943 & 0.889 & 0.933 & \bf \underline{0.957} & 0.936 & 0.936 \\ \hline
    \bf Average ($r$) & 0.905 & 0.940 & 0.904 & 0.942 & \bf \underline{0.952} & \bf 0.949 & \bf 0.949\\ \hline
  \end{tabular}
  \end{adjustbox}
  \caption{Pearson correlations on the WMT17 metrics task. The best correlation of each language pair is underlined, while correlations of our methods that outperform or match any other method are in bold. 
  }
  \label{tab:mt_word_best}
\end{table*}

Pearson ($r$) correlations with human evaluation are presented in Table~\ref{tab:mt_word_best}. Our family of metrics outperforms all the rest in several language pair translations. Moreover, our metrics show the highest correlation to human evaluation when considering the average correlation across all language pairs. BERTScore results were calculated using the embeddings from the last fifth layer of the aforementioned BERT model. 

\subsection{Text summarization}
\label{sec:text_summarization}

We further assessed text summarization with the TAC-2009 dataset\footnote{http://tac.nist.gov/}, consisting of news articles from ten different topics, with four reference summaries and fifty-five evaluation summaries from summarization systems per article. We evaluate each evaluation summary independently, performing a summary-level evaluation. Two scores were assigned to each evaluation summary: the \textit{pyramid score}, which evaluates the semantic similarity between the reference and evaluation summaries, and the \textit{responsiveness score}, that measures the overall quality of the evaluation summary in terms of grammar and content.


Table~\ref{tab:human_eval_text_summarization} shows the Kendall ($k$), Pearson ($r$), and Spearman ($p$) correlation to human evaluation for each score. Considering Kendall and Spearman correlations, our family of methods outperforms all the rest on responsiveness score. Moreover, ME$_\text{Petersen}$ and ME$_\text{CAPTURE}$ outperform all methods on the above correlations on the pyramid score. 
Considering Pearson correlation, our family of methods outperforms PRD, and at least one of our metrics consistently outperforms IMPAR on both scores. We hypothesize that the lower Pearson correlations of our metrics could be explained by the instability of the population estimation process due to the low amount of samples, \textit{i.e. }reference and evaluation words. 

\begin{table} 
\vskip 0.15in
\begin{center}
\begin{small}
\begin{sc}
\centering
\begin{tabular}{ccccccccc}
\toprule
\bf Metrics & BERTScore & MoverScore & PRD & IMPAR & \bf ME$_\text{Schnabel}$ & \bf ME$_\text{Petersen}$ & \bf ME$_\text{CAPTURE}$\\
\midrule
\bf Responsiveness ($k$) & - & 0.482 & 0.398 & 0.481 & \textbf{0.483} & \underline{\textbf{0.487}} & \textbf{0.484} \\
\bf Responsiveness ($r$) & 0.739 & \underline{0.754} & 0.564 & 0.743 & 0.739 & 0.683 & 0.747\\
\bf Responsiveness ($p$) & 0.580 & 0.594 & 0.501 & 0.594 & \textbf{0.595} & \underline{\textbf{0.598}} & \textbf{0.596} \\
\midrule
\bf Pyramid ($k$) & - & 0.550 & 0.444 & 0.541 & 0.548 & \textbf{0.555} & \underline{\textbf{0.565}}\\
\bf Pyramid ($r$) & 0.823 & \underline{0.831} & 0.658 & 0.804 & 0.813 & 0.770 & 0.808 \\
\bf Pyramid ($p$) & 0.703 & 0.701 & 0.588 & 0.693 & 0.698 & \textbf{0.704} & \underline{\textbf{0.718}}\\
\bottomrule
\end{tabular}
\end{sc}
\end{small}
\end{center}
\caption{Summary-level correlations to human evaluation on TAC 2009. Bold values represent correlations from our proposed metrics that outperform or match all the other methods. The best correlations of each score are underlined. BERTScore results were taken from~\newcite{moverscore}.
}
\label{tab:human_eval_text_summarization}
\end{table}


\section{Conclusion}

In this work, we present a family of methods derived from popular population size estimators that have been widely used in ecology in the past several decades. We show that our family of methods is able to assess language systems under different representations effectively, \textit{i.e.} using contextualized word and sentence embeddings. Our methods show a high correlation to human evaluation on challenging language generation tasks as well as the desired sensitivity to detect mode collapse and quality detriment.

In the future, we would like to evaluate our family of metrics on image generation tasks, reinforcing the general applicability of our methods. Moreover, we plan to extend our family of methods to also cover popular open populations estimation methods, where the population size may vary over time. In the end, we hope that combining the information from closed and open population methods will improve the overall assessment of language systems, further fostering the adoption of ecology methods in NLP.

\bibliographystyle{coling}
\bibliography{coling2020}

\newpage
\appendix

\section{Theoretical discussions}

We will briefly study the validity of our methods when assessing two equal sets, \textit{i.e.} when the reference set is identical to the evaluation set. Formally, we define that: 

\theoremstyle{definition}
\begin{definition}{}
    Two sets $S$ and $S'$ are equal if $S \subseteq S'$ and $S' \subseteq S$.
\end{definition}

\begin{theorem}
\label{theo:petersen}
    Considering two equal sets $S$ and $S'$, ME$_\text{Petersen}(S,S')$ returns its maximum score of 1.
\end{theorem}

\begin{proof}

Since $\sum\limits_{s' \in S'} f(s', S) = \sum\limits_{s \in S} f(s, S') = |S|$, following Equation~\ref{eq:petersen} we have:

\begin{equation}
    \widehat{P}_\text{Petersen}(S,S') = \dfrac{2|S|\times2|S|}{2|S|} = 2|S|.
\end{equation}

Since $P = 2|S|$, following Equation~\ref{eq:accuracy} we have:

\begin{equation}
A(P,\widehat{P}_\text{Petersen}(S,S')) = A(2|S|,2|S|) = 0.
\end{equation}

Finally, adopting Equation~\ref{eq:mark_evaluate}, we conclude the proof: 

\begin{equation}
    \text{ME}_\text{Petersen}(S,S') = 1 - A(P,\widehat{P}_\text{Petersen}(S,S')) = 1 - 0 = 1.
\end{equation}

\end{proof}

\begin{theorem}
\label{theo:schnabel}
    Considering two equal sets $S$ and $S'$, ME$_\text{Schnabel}(S,S')$ returns its maximum score of 1.
\end{theorem}

\begin{proof}

Since $\sum\limits_{s' \in S'} \sum\limits_{s \in S} f(s, \text{NN}_k(s', S')) = |S|\times(K+1)$ and $\sum\limits_{i = 1}^{|S'|} \sum\limits_{j = 1}^{|S|} \Big(f(s_j, \text{NN}_k(s'_i, S')) + |M(i,S,S') \cap \text{NN}_k(s'_i, S')|\Big) = |S'|\times\Big((K+1)+(K+1)\Big)$, following Equation~\ref{eq:schnabel} we have:

\begin{equation}
\begin{split}
\widehat{P}_\text{Schnabel}(S, S') = \dfrac{\Big((K+1)\times|S'| + |S|\times(K+1)\Big) \times 2|S|}{|S'|\times\Big((K+1) + (K+1)\Big)}
= \dfrac{\Big(2\times|S|\times(K+1)\Big)\times
2|S|}{\Big(2\times|S|\times(K+1)\Big)}
= 2|S|.
\end{split}
\end{equation}

Since $P = 2|S|$, following Equation~\ref{eq:accuracy} we have:

\begin{equation}
A(P,\widehat{P}_\text{Schnabel}(S,S')) = A(2|S|,2|S|) = 0.
\end{equation}

Finally, adopting Equation~\ref{eq:mark_evaluate}, we conclude the proof: 

\begin{equation}
    \text{ME}_\text{Schnabel}(S,S') = 1 - A(P,\widehat{P}_\text{Schnabel}(S,S')) = 1 - 0 = 1.
\end{equation}

\end{proof}

\begin{theorem}
\label{theo:capture}
    Considering two equal sets $S$ and $S'$, ME$_\text{CAPTURE}(S,S')$ returns its maximum score of 1.
\end{theorem}

\begin{proof}

Since $\sum\limits_{s \in S} \sum\limits_{s' \in S'} \Big( f(s', \text{NN}_k(s, S)) + |\text{NN}_k(s, S)| \Big) = \sum\limits_{s' \in S'} \sum\limits_{s \in S} \Big( f(s, \text{NN}_k(s', S')) + |\text{NN}_k(s', S')|\Big) = 2|S|(K+1)$, following Equation~\ref{eq:capture} we have:

\begin{equation}
\begin{split}
    L_n(P_\text{CAPTURE};S,S') = \ln\Big(\dfrac{P_\text{CAPTURE}!}{(P_\text{CAPTURE}-2|S|)!}\Big) + 4|S|(K+1) \times ln(4|S|(K+1)) + (2|S|\\P_\text{CAPTURE}-4|S|(K+1)) \times \ln(2|S|P_\text{CAPTURE}-4|S|(K+1)) - (2|S|P_\text{CAPTURE})\ln(2|S|P_\text{CAPTURE}).\\
\end{split}
\end{equation}

Iterating through $P_\text{CAPTURE} \in \mathbb{N}_{\geq 2|S|}$ and substituting $|S|=10$ and $K=1$ as an illustration, we have:

\begin{equation}
\begin{split}
    L_N(2|S|; S,S') \approx -158 \quad and \quad L_N(2|S|+1; S,S') \approx -159.
\end{split}
\end{equation}

Since $L_n(2|S|; S,S') > L_n(2|S|+1; S,S'), P_\text{CAPTURE} = 2|S|$ maximizes the likelihood function and, following Equation~\ref{eq:capture_2}, $\widehat{P}_\text{CAPTURE} = 2|S|$.

Since $P = 2|S|$, following Equation~\ref{eq:accuracy} we have:

\begin{equation}
A(P,\widehat{P}_\text{CAPTURE}(S,S')) = A(2|S|,2|S|) = 0.
\end{equation}

Finally, adopting Equation~\ref{eq:mark_evaluate}, we conclude the proof: 

\begin{equation}
    \text{ME}_\text{CAPTURE}(S,S') = 1 - A(P,\widehat{P}_\text{CAPTURE}(S,S')) = 1 - 0 = 1.
\end{equation}

\end{proof}

\section{Additional experiments on dialogue generation}

Human evaluation correlation on assessing language generation with default $K$, as well as a comparison with InferSent and SBERT embeddings from BERT-base and BERT-large, are provided in Table~\ref{tab:human_eval_correlations_default}.
We observe that the relative performance between all methods does not change when compared to using the best $K$, with our family of methods showing the overall best performance between the compared methods and embeddings. Furthermore, SBERT-based embeddings tend to show a higher correlation than InferSent embeddings across all correlations and methods, with the exception of IMPAR's $r$ and $k$. This goes in accordance with several recent works that show that contextualized embeddings from BERT seem to help across a wide variety of tasks~\cite{liu_linguistic,li_deep,gabriel2019cooperative,bertr,yoshimura_filtering}.

\begin{table}[b]
\vskip 0.15in
\begin{center}
\begin{small}
\begin{sc}
\centering
\begin{adjustbox}{width=1.0\textwidth}
\begin{tabular}{cccccccccccccccccccc}
\toprule
& \multicolumn{3}{c}{FID} & \multicolumn{3}{c}{PRD} & \multicolumn{3}{c}{IMPAR} & \multicolumn{3}{c}{\bf ME$_\text{Schnabel}$} & \multicolumn{3}{c}{\bf ME$_\text{Petersen}$} & \multicolumn{3}{c}{\bf ME$_\text{CAPTURE}$}\\
\cmidrule(lr){2-4}\cmidrule(lr){5-7}\cmidrule(lr){8-10}\cmidrule(lr){11-13}\cmidrule(lr){14-16}\cmidrule(lr){17-19}
\textbf{Corr.} & {ISENT} & {sSBERT} & {SBERT} & {ISENT} & {sSBERT} & {SBERT} & {ISENT} & {sSBERT} & {SBERT} & {ISENT} & {sSBERT} & {SBERT} & {ISENT} & {sSBERT} & {SBERT} & {ISENT} & {sSBERT} & {SBERT} \\
\midrule
$r$ & 0.838 & 0.860 & 0.902 & 0.669  & 0.661 & 0.684  & 0.708  & 0.629 & 0.633  & \bf 0.905  & \bf 0.883  & \bf 0.917  & 0.824  & 0.828 & 0.872  & \bf 0.903  & \bf 0.882 & \bf 0.902 \\
$k$ & 0.511 & 0.867 & 0.867 & 0.689  & 0.778 & 0.822  & 0.689  & 0.467 & 0.600  & \bf 0.778  & \bf \underline{0.911} & \bf \underline{0.911} & \bf \underline{0.867}  & \bf \underline{0.911} & \bf \underline{0.911}  & \bf \underline{0.867}  & \bf \underline{0.911} & \bf 0.867 \\
$p$ & 0.733 & 0.964 & 0.964 & 0.830  & 0.903 & 0.939  & 0.879  & 0.552 & 0.903  & \bf 0.903  & \bf \underline{0.976} & \bf \underline{0.976} & \bf \underline{0.964}  & \bf \underline{0.976} & \bf \underline{0.976}  & \bf \underline{0.964}  & \bf \underline{0.976} & \bf 0.964\\
\bottomrule
\end{tabular}
\end{adjustbox}
\end{sc}
\end{small}
\end{center}
\caption{Correlations to human evaluation regarding the fluency of 10 different models with default $K$. ISENT refers to InferSent embeddings, sSBERT refers to sentence embeddings from BERT-base ('bert-base-nli-mean-tokens'), and SBERT refers to sentence embeddings from BERT-large ('bert-large-nli-mean-tokens'). For each embedding type, best scores of each correlation are underlined, while bold values represent the correlations where our methods outperform or match all of the other methods' performance. Absolute correlation values are presented for FID.}
\label{tab:human_eval_correlations_default}
\end{table}

\section{Additional experiments on machine translation}

We further experimented with assessing machine translation systems using contextualized sentence embeddings. To achieve this, we use all the reference translations as reference samples and the translations of each translation system as evaluation samples. We perform this assessment individually for each translation system available for each language pair.

Pearson ($r$) correlations are presented in Table~\ref{tab:mt_sentence_default}. Considering the average across all language pairs, our family of methods outperforms PRD and IMPAR. Note that, as expected, using contextualized sentence embeddings shows lower performance than contextualized word embeddings (Table~\ref{tab:mt_word_default}) due to the finer-granularity of the assessment in the latter case.

\begin{table*}
  \centering
  \begin{tabular}{ccccccccccccccc}
    \hline
    \textbf{Translations} & PRD & IMPAR & \textbf{ME$_\text{Schnabel}$} & \textbf{ME$_\text{Petersen}$} & \textbf{ME$_\text{CAPTURE}$}\\
    \hline
    \bf cs-en ($r$) & 0.979 & 0.912 & \bf \underline{0.993} & \bf 0.989 & 0.971\\ \hline
    \bf de-en ($r$)  & \underline{0.885} & 0.794 & 0.869 & 0.857 & 0.845\\ \hline
    \bf ru-en ($r$) & 0.931 & 0.897 & \bf \underline{0.949} & 0.904 & 0.842\\ \hline
    \bf tr-en ($r$) & 0.833 & 0.866 & 0.857 & \bf 0.874 & \textbf{\underline{0.922}}\\ \hline
    \bf zh-en ($r$) & 0.582 & 0.727 & \textbf{0.745} & \bf 0.735 & \textbf{\underline{0.839}}\\ \hline
    \bf Average ($r$) & 0.842 & 0.839 & \bf \underline{0.883} & \bf 0.872 & \bf 0.850\\ \hline
  \end{tabular}
  \caption{Pearson correlations for the WMT17 metrics task using contextualized sentence SBERT embeddings (\textit{'bert-base-nli-max-tokens'}) with default $K$. The best correlation of each language pair is underlined. Correlations where our methods outperform or match all of the other methods are highlighted in bold.}
    \label{tab:mt_sentence_default}
\end{table*}

\begin{table*}
  \centering
  \begin{tabular}{ccccccccccccccc}
    \hline
    \textbf{Translations} & PRD & IMPAR & \textbf{ME$_\text{Schnabel}$} & \textbf{ME$_\text{Petersen}$} & \textbf{ME$_\text{CAPTURE}$}\\
    \hline
    \bf cs-en ($r$) & \underline{0.992} & 0.981 & 0.975 & 0.970 & 0.971\\ \hline
    \bf de-en ($r$) & 0.769 & 0.934 & \bf 0.940 & \bf \underline{0.953} & \bf \underline{0.953}\\ \hline
    \bf ru-en ($r$) & \underline{0.933} & 0.888 & 0.887 & 0.895 & 0.897\\ \hline
    \bf tr-en ($r$) & 0.935 & \underline{0.941} & 0.917 & 0.917 & 0.929\\ \hline
    \bf zh-en ($r$) & 0.889 & 0.895 & \bf \underline{0.957} & \bf 0.936 & \bf 0.936\\ \hline
    \bf Average ($r$) & 0.904 & 0.928 & \bf 0.935 & \bf 0.934 & \bf \underline{0.937}\\ \hline
  \end{tabular}
  \caption{Pearson correlations for the WMT17 metrics task using contextualized word embeddings with default $K$. The best correlation of each language pair is underlined. Correlations where our methods outperform or match all of the other methods are highlighted in bold. }
  \label{tab:mt_word_default}
\end{table*}

\end{document}